\documentclass[10pt,twocolumn,letterpaper]{article}

 \usepackage[pagenumbers]{cvpr} % To force page numbers, e.g. for an arXiv version

\usepackage{graphicx}
\usepackage{parskip}
\definecolor{cvprblue}{rgb}{0.21,0.49,0.74}
\usepackage[pagebackref,breaklinks,colorlinks,allcolors=cvprblue]{hyperref}

\def\paperID{5} % *** Enter the Paper ID here
\def\confName{CVPR}
\def\confYear{2025}

\title{Uncovering Branch specialization in InceptionV1 using k sparse autoencoders}

\author{Matthew Bozoukov\\
Miramar college\\
{\tt\small matthewbozoukov123@gmail.com}}
\begin{document}
\maketitle
\begin{abstract}
Sparse Autoencoders (SAEs) have shown to find interpretable features in neural networks from polysemantic neurons caused by superposition. Previous work has shown SAEs are an effective tool to extract interpretable features from the early layers of InceptionV1. Since then, there have been many improvements to SAEs but branch specialization is still an enigma in the later layers of InceptionV1. We show various examples of branch specialization occuring in each layer of the mixed4a-4e branch, in the 5x5 branch and in one 1x1 branch. We also provide evidence to claim that branch specialization seems to be consistent across layers, similar features across the model will be localized in the same convolution size branches in their respective layer.
\end{abstract}    
\section{Introduction}
\label{sec:intro}

Previous work \cite{livgorton} uses SAEs(sparse autoencoders)\cite{cunningham2023sparseautoencodershighlyinterpretable} to uncover interpretable features from the early layers of InceptionV1 \cite{szegedy2014goingdeeperconvolutions} to address the main roadblock for \cite{olah2020an} that polysemantic neurons imposed on the analysis of early mechanistic interpretability. The superposition\cite{elhage2022toymodelssuperposition} hypothesis gives a potential reason why this happens, where combinations of neurons try to encapsulate a feature's representation. Sparse autoencoders\cite{cunningham2023sparseautoencodershighlyinterpretable}, a variant of dictionary learning, work well to disentangle these combinations of neurons. \cite{livgorton, mixed5bBlogPost} Show that SAE's help find previously missing curve detectors and are able to find interpretable features from polysemantic neurons in the mixed5b layer. However one question remained from the original project: What kind of branch specialization\cite{voss2021branch} occurs in the later layers of InceptionV1?

This question is hard to answer with vanilla SAE's because they suffer from "dead latents"\cite{gao2024scalingevaluatingsparseautoencoders,templeton2024scaling}, features that do not activate often, if at all. \cite{templeton2024scaling} shows that a 34 million feature vanilla autoencoder, only 12 million features are alive and this number can decrease to where up to 90 percent of features never activate. For layers like mixed4a-4e, where superposition seems to be prevalent, having many dead features will make the search for branch specialization

difficult.\cite{gao2024scalingevaluatingsparseautoencoders} k-sparse autoencoders and using the technique of tying the encoder weights and decoder weights together, enable the autoencoder to decrease the number of dead features while also decreasing the reconstruction error.

Using k-sparse autoencoders plus the techniques of \cite{gao2024scalingevaluatingsparseautoencoders}, we show the branch specialization of every 5x5 convolution branch from layers mixed4a-4e and the 1x1 conv branch of mixed4a and mixed4d. We also show that branch specialization forms features progressively. For example, mixed4b-4e starts from animal parts in specific positions to general animal parts to specific animals. We prove this through circuit analysis\cite{olah2020zoom}. We also show through an 2d UMAP projection how the decoder vectors of the mixed4b 5x5 branch and mixed4c 5x5 branch, that they are very similar in how they organize themselves, indicating that are related in some way.

\section{Setup and Related work}
\label{sec:formatting}

 We formally define an k-sparse autoencoder. We also define the dimension of our input space to be d and the latent dimension be l.
 \begin{equation}
     z=TopK(W_{enc}(x-b_{decoded})+b_{encoded}
 \end{equation}

\begin{equation}
    x'=W_{dec}z+b_{decoded}
\end{equation}

where x $\in$ $\mathbb{R}^{d}$, $W_{enc} \in \mathbb{R}^{d \times l}, W_{dec} \in \mathbb{R}^{ l \times d}$ $b_{enc} \in \mathbb{R}^{l}$,$b_{dec} \in \mathbb{R}^{d}$

The TopK activation function takes the top k elements in the vector and zeroes out the rest. The advantage of this is that we set how many features we want active, removing the need of a sparsity penalty. We chose a k value of 32, for every SAE we trained. The loss is an MSE error with the original input and the reconstruction of the SAE.

To avoid dead features we tie the weights of the encoder matrix and the decoder matrix, essentially making the decoder matrix the transpose of the encoder matrix. We use an expansion factor of 16 for the latent space for all the SAEs we train.
\begin{figure*}[!t]
  \centering
   
  \includegraphics[width=1\textwidth, height=1\textheight, keepaspectratio,scale=1]{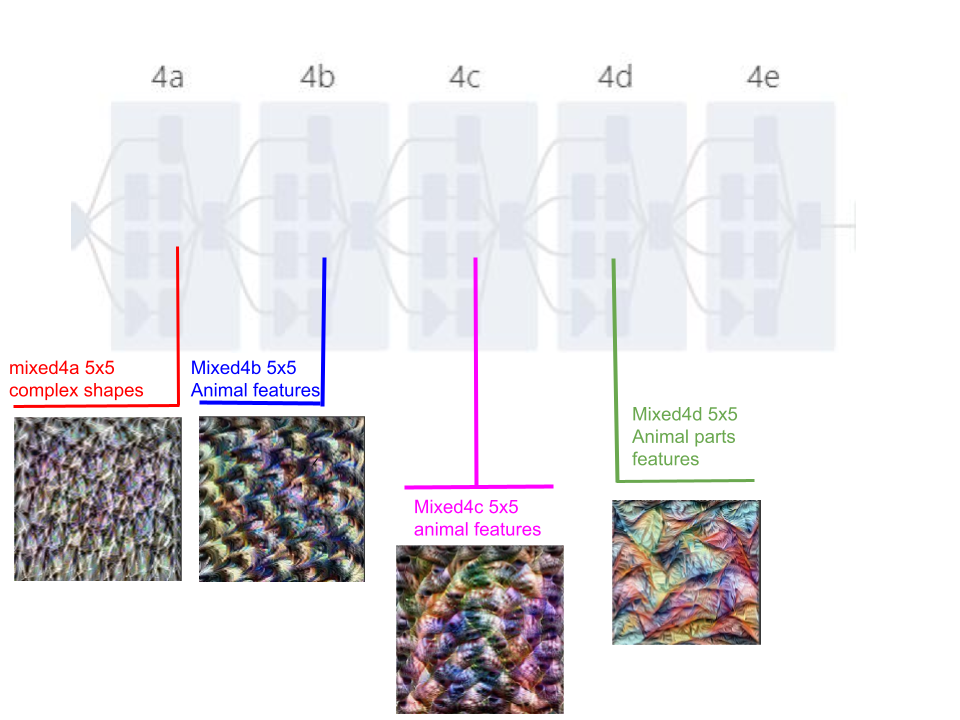}
  \caption{This figure shows the various branch specializations across the layers mixed4a-mixed4e. We observe that the 5x5 branch of each layer (besides the mixed4a layer) are something animal related. The mixed4b layer primarily focuses on specific orientations and poses of animals. Mixed4c is kind of a mixed bag of features, some orientation/poses, some general animal features. Mixed4d and 4e(not depicted) seem to be species specific features.  }
  \label{fig:neurons}
\end{figure*}

To train we used the ImageNet \cite{5206848} dataset and for each image sampled 10 activation vectors that had the highest norm in the image. An activation vector, in the context of CNNs, would be a grid position in the output activation. We train on right around a billion activation vectors per SAE. Taking the highest normed activation vectors allows us to save on compute, since the majority of activation vectors are of backgrounds.
\subsection{Circuit analysis}
\cite{olah2020zoom} defines a circuit as a "A subgraph of a neural network. Nodes correspond to neurons or directions (linear combinations of neurons). Two nodes have an edge between them if they are in adjacent layers. The edges have weights which are the weights between those neurons (or $n_1Wn_2^T$ if the nodes are linear combinations)" 

Features can be replaced instead of neurons in the expression above to find connections between them. However we can also perform circuit analysis by finding the feature's most related activation vector, ablating it, then finding the activation vector in the next layer that is affected by this ablation the most, and then finding the most related feature through cosine similarity.\cite{olah2020zoom} form preliminary circuits but were limited from their effectiveness in the later layers, due to the number of polysemantic neurons. \cite{mixed5bBlogPost} shows the effectiveness of using sparse autoencoders to find circuits. They also found a circuit from mixed5a-mixed5b. Circuit analysis can be used to show how features are connected, but also can be used to show how information is shared throughout the model.
\begin{figure*}[ht]
  \centering
   
  \includegraphics[width=1\textwidth, height=1\textheight, keepaspectratio,scale=1]{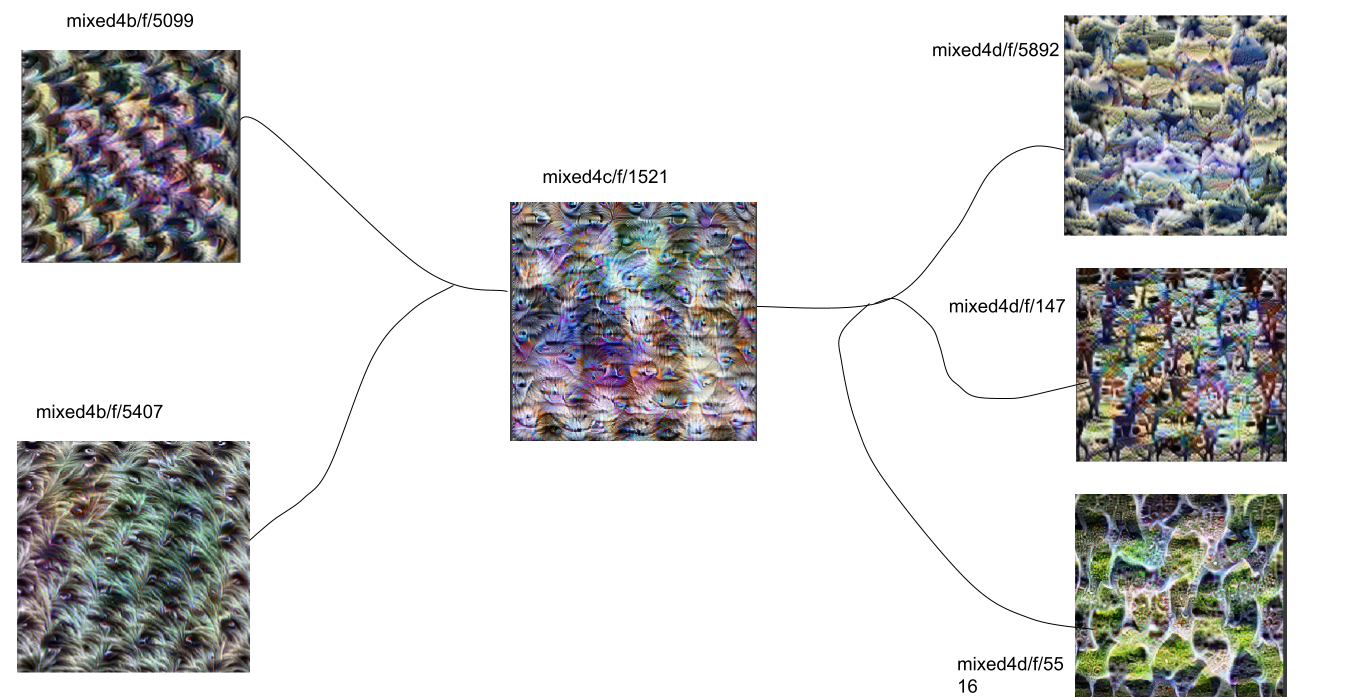}
  \caption{This circuit is an example of how the 5x5 branches of each layer seem to have circuits connecting one another. The mixed4b features are localized to the 5x5 branch primarily and are of animals orientated facing left, and facing right. The feature in layer Mixed4c that they both produce is a feature that detects faces of animals. Again, this feature has the largest neuron contribution from neurons in the 5x5 branch. The last part of the circuit shows the last feature spreading out into a feature that detects dogs with fluffy white fur, and two features that seem to detect dog legs. These features have the largest neuron contribution from neurons in the Mixed4d 5x5 branch.}
  \label{fig:circuit}
\end{figure*}

%-----------------------------------------------------------------------
\subsection{Branch-specific SAE's}
Inceptionv1's\cite{szegedy2014goingdeeperconvolutions} later layer's consist of parallel convolution branches. At the end of the layer their outputs are concatenated. \cite{livgorton} notices that cross branch superposition is significant and it would be better to train an SAE across the entire layer rather than train on specific branches. We train on the entire layer and features found using layer SAEs will simply be denoted as layer/f/number. Neurons will be just be labeled as layer/n/number.
%-------------------------------------------------------------------------

\subsection{Analysis Methods}
We use feature visualization{\cite{olah2017feature}} Feature visualization helps us isolate what causes a specific feature to activate, in specific how it activates out of distribution.

 The mixed4 layer feature visualizations, however, tend to be very abstract so it can be hard to interpret them. However by looking at dataset examples in distribution, we can get a better sense of what these feature visualizations are trying to model. More often than not the visualizations are consistent with the dataset examples.

\subsection{Feature visualization and Dataset examples}

We use the Lucent library \cite{greentfrapp2025lucent} for feature visualization and for data set examples, we look for images in the imagenet\cite{5206848} dataset that activate a feature or neuron in a range of various activation levels.

%-------------------------------------------------------------------------

\begin{figure}[ht]
  \centering
   
  \includegraphics[width=.5\textwidth, height=1\textheight, keepaspectratio,scale=1]{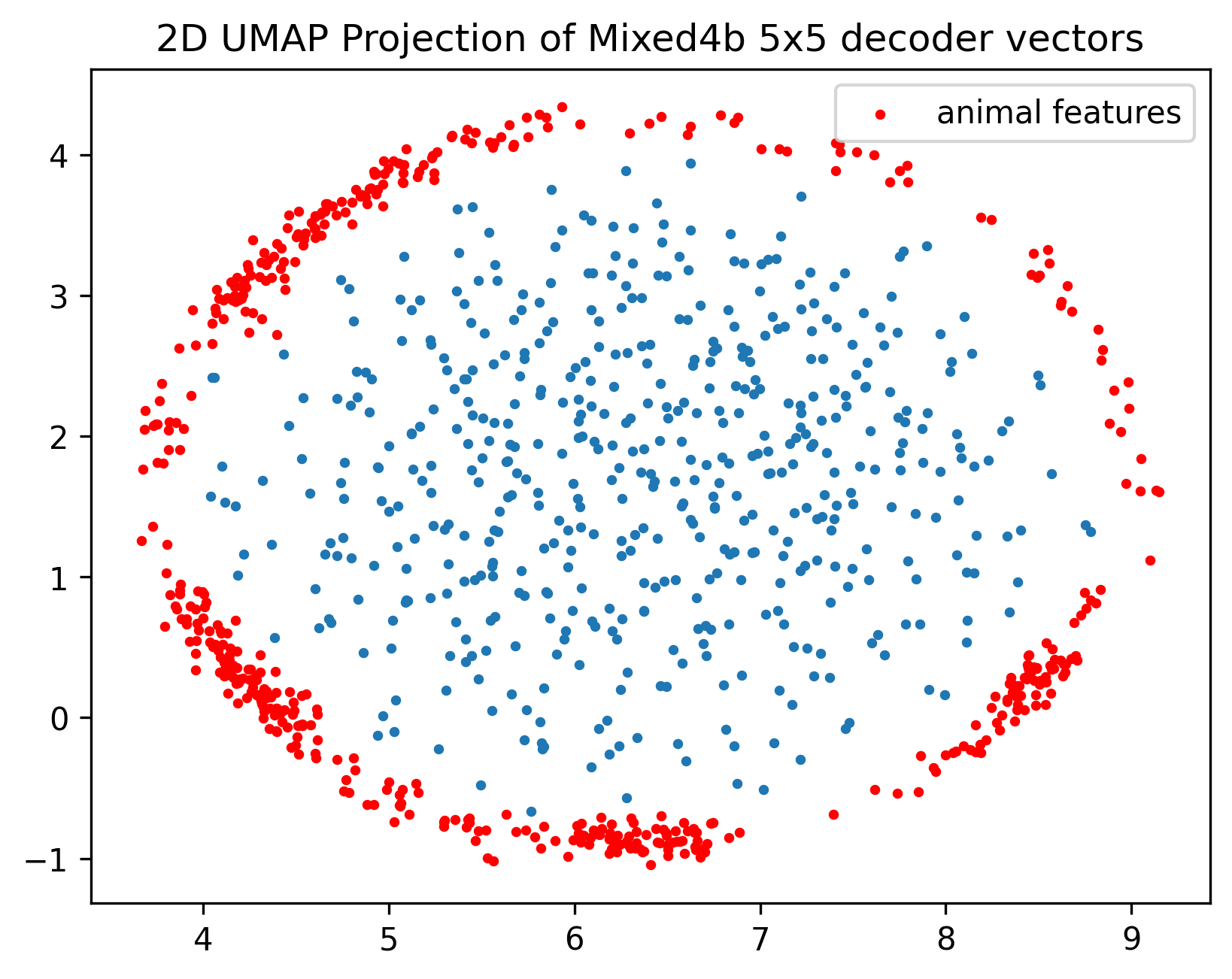}
  \caption{A UMap projection of the decoder vectors of Mixed4b 5x5. Around all the features the animal specific features seem to form a manifold. }
  \label{fig:branch1}
\end{figure}
\begin{figure}[ht]
  \centering
   
  \includegraphics[width=.5\textwidth, height=1\textheight, keepaspectratio,scale=1]{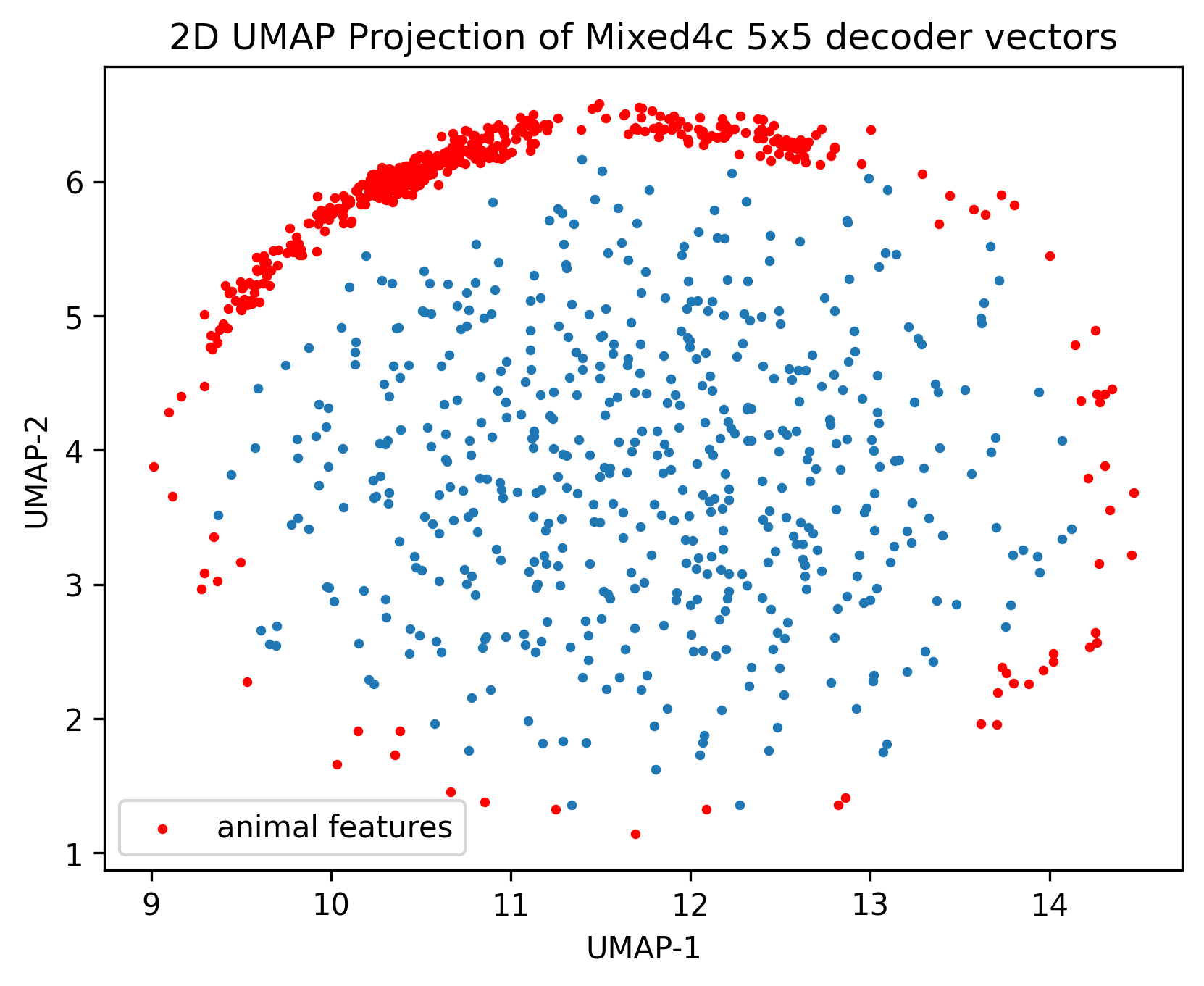}
  \caption{ A UMap projection of the decoder vectors of Mixed4c 5x5.Around all the features the animal specific features seem to form a manifold.}
  \label{fig:branch}
\end{figure}
\section{Results}
    \cite{livgorton} was an early report of how one could use SAE's for a network like InceptionV1. We explore deeper into the phenomena and features in the fourth layer of InceptionV1. While we do not provide a full \textit{taxomony} of all the features in the branches we provide enough examples to show a general trend of branch specialization. Our results are as follows:

\begin{itemize}
    \item We that find branch specialization in layers mixed4a-mixed4e, which includes specialization in animals in specific poses/animals in specific spots in an image features, color specific features, animal body parts, and dog species specific features.

    \item We that find branch specialization seems to be similar across layers, meaning that the same kind of features seem to arise from the same convolution sizes but across different layers. The evidence suggesting these are circuits formed specifically consisting of features localized from the same type of convolution branch and Umap projections of Mixed4b's and Mixed4c's decoder vectors.

\end{itemize}

\subsection{Identifying Branch Specialization}
\cite{livgorton,voss2021branch} show early signs of the emergent behavior of Branches to specialize in certain features. \cref{fig:neurons} gives 5 examples of specialization behavior in Mixed4a-4e. \cite{voss2021branch} hypothesize that mixed4a's 5x5 branch specialized in complex geometric shapes. This hypothesis seems to be true from the features we found. 

One interesting thing we found is that the mixed4b 5x5 branch and mixed4c 5x5 branch seem to have the same kind of specialization: specific orientation or position of animals. For example, these orientation features of the 5x5 branch seem to only detect animals which are on the left side of the picture or the right side or animals that are facing left or facing right.

Additional evidence is shown in Appendix A in the form of graphs showing the localization of features in each branch.

\subsection{Branch Specialization progression}

\cite{voss2021branch} hypothesis for why branch specialization occurs is due to the nature of training these networks: the first half of the branch will want to form a prior for the second half, and in turn the second half of the branch will have similar features to the first half. Could we extend this to branches of different layers? For example, are the specialized features of the mixed4c 5x5 branch similar to the specialized features of the mixed4b 5x5 branch? The answer seems to be yes. We show this in two ways, through circuit analysis and through dimensionality reduction.

We can see from \cref{fig:circuit} that this circuit starts with features that detect animals oriented left and right in Mixed4b. These features are connected through a circuit that form an animal feature, but in specific animals positioned near the bottom of the image feature. This feature also seems to be around 40\% of its total norm, consisting of neurons in the 5x5 branch, meaning that it is specialized in the 5x5 branch. Then this feature seems to separate into various dog-specific features(fur and lower body). We also notice that Mixed4b and Mixed4c seem to detect very similar features. These kind of features shared by Mixed4b and Mixed4c seem to be related by a copy gate similar to how \cite{elhage2021mathematical} language transformers have a "copy gate".

We also show a 2d Umap projection of both the mixed4b and mixed4c 5x5 decoder vectors in \cref{fig:branch1} and \cref{fig:branch}. These diagrams show how the features of mixed4b and mixed4c form similar rings or manifolds around all the other features in the same layer. This shows similar structure in the mixed4b and mixed4c layers except for the apparent clustering of features near the top in \cref{fig:branch}. This differs from the somewhat uniform distribution features in \cref{fig:branch1}.

\section{Conclusion}

While this is great progress on understanding branch specialization and showing examples of cross layer superposition, there is still more work to be done in understanding InceptionV1 on a zoomed out scale. This can be done through an automated interpretability approach \cite{OpenAI2023Neurons,shaham2025multimodalautomatedinterpretabilityagent}. Another important step would be to create dashboards to visualize all features at once\cite{OpenAI2021Microscope}.

{
    \small
    \bibliographystyle{ieeenat_fullname}
    \bibliography{main}
}

\newpage
\appendix
\onecolumn

\section{Graphs for Branch specialization}

\begin{figure*}[ht]
  \centering
   
  \includegraphics[width=.75\textwidth, height=1\textheight, keepaspectratio,scale=1]{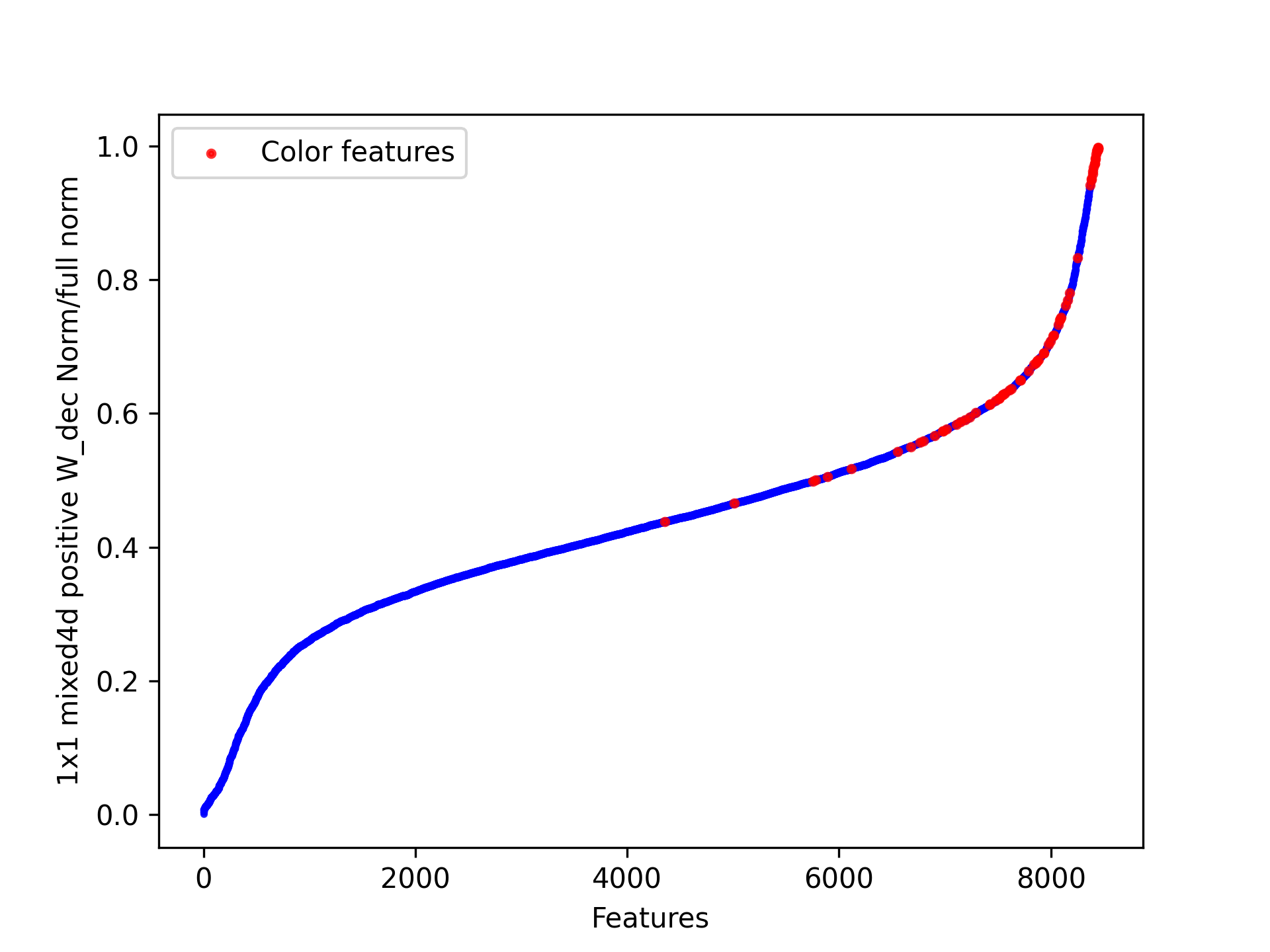}
  \caption{ How much a learned feature from the sparse autoencoder trained on all of mixed4d is represented by the neurons in branch 1x1. We can see that color features generally are represented in a large portion by the neruons in the 1x1 branch}
  \label{fig:graph1}
\end{figure*}
\begin{figure*}[ht]
  \centering
   
  \includegraphics[width=.75\textwidth, height=1\textheight, keepaspectratio,scale=1]{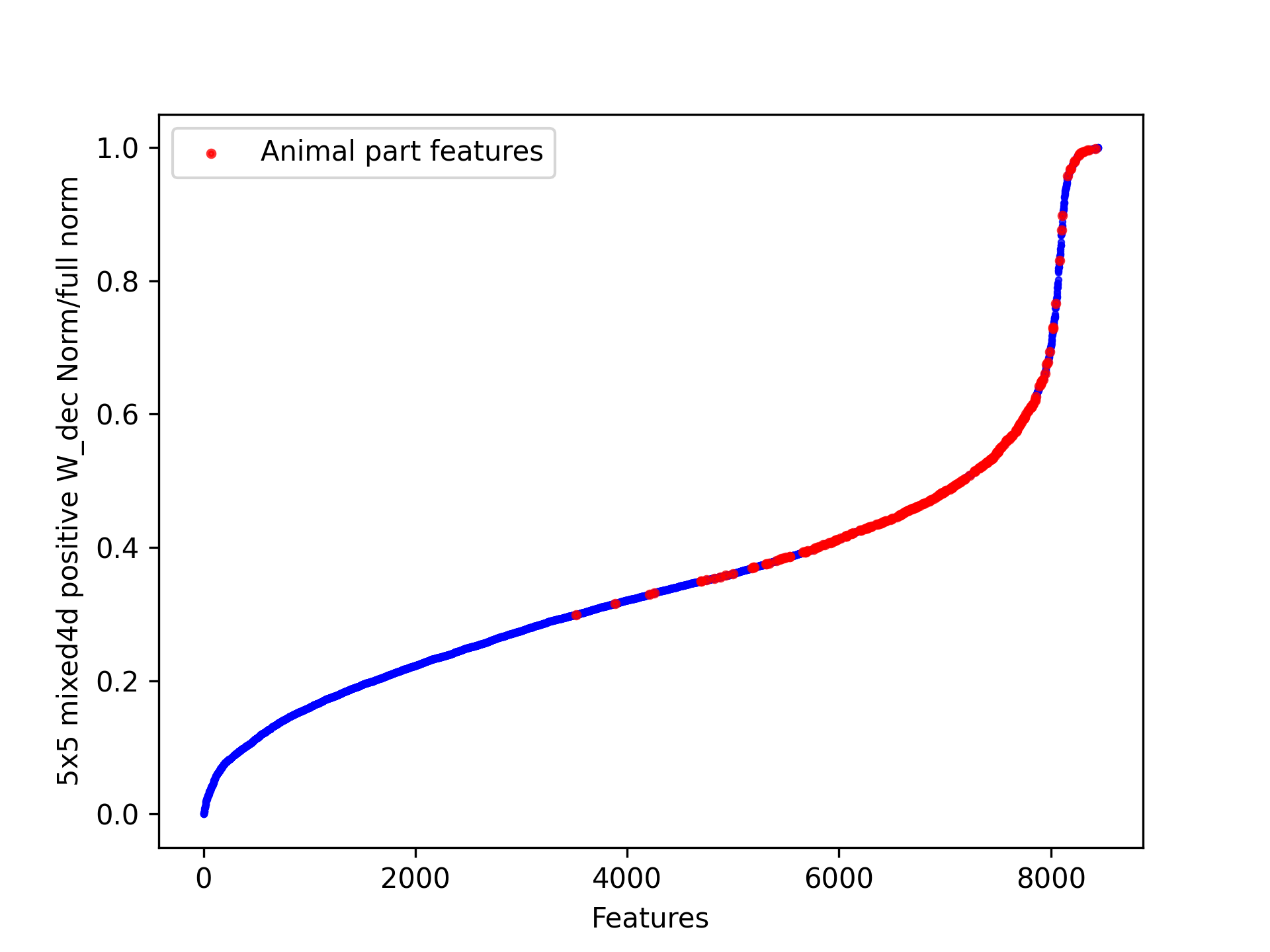}
  \caption{ How much a learned feature from the sparse autoencoder trained on all of mixed4d is represented by the neurons in branch 5x5. The animal part features seem to be more uniform between .4 and .6, still indicating that a majority of animal features are represented by the 5x5 neurons}
  \label{fig:graph2}
\end{figure*}
\begin{figure*}[ht]
  \centering
   
  \includegraphics[width=1\textwidth, height=1\textheight, keepaspectratio,scale=1]{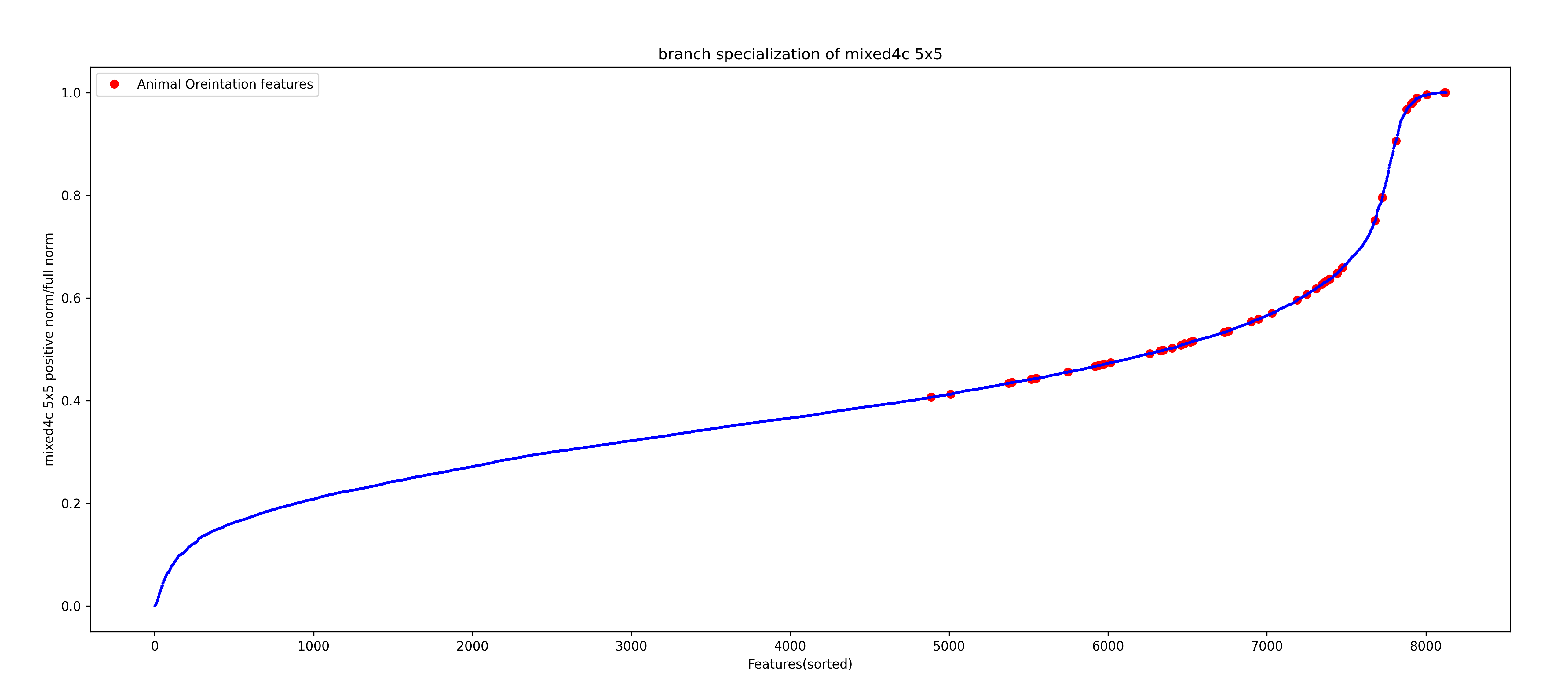}
  \caption{ How much a learned feature from the sparse autoencoder trained on all of mixed4c is represented by the neurons in branch 5x5.}
  \label{fig:graph2}
\end{figure*}
\begin{figure*}[ht]
  \centering
   
  \includegraphics[width=1\textwidth, height=1\textheight, keepaspectratio,scale=1]{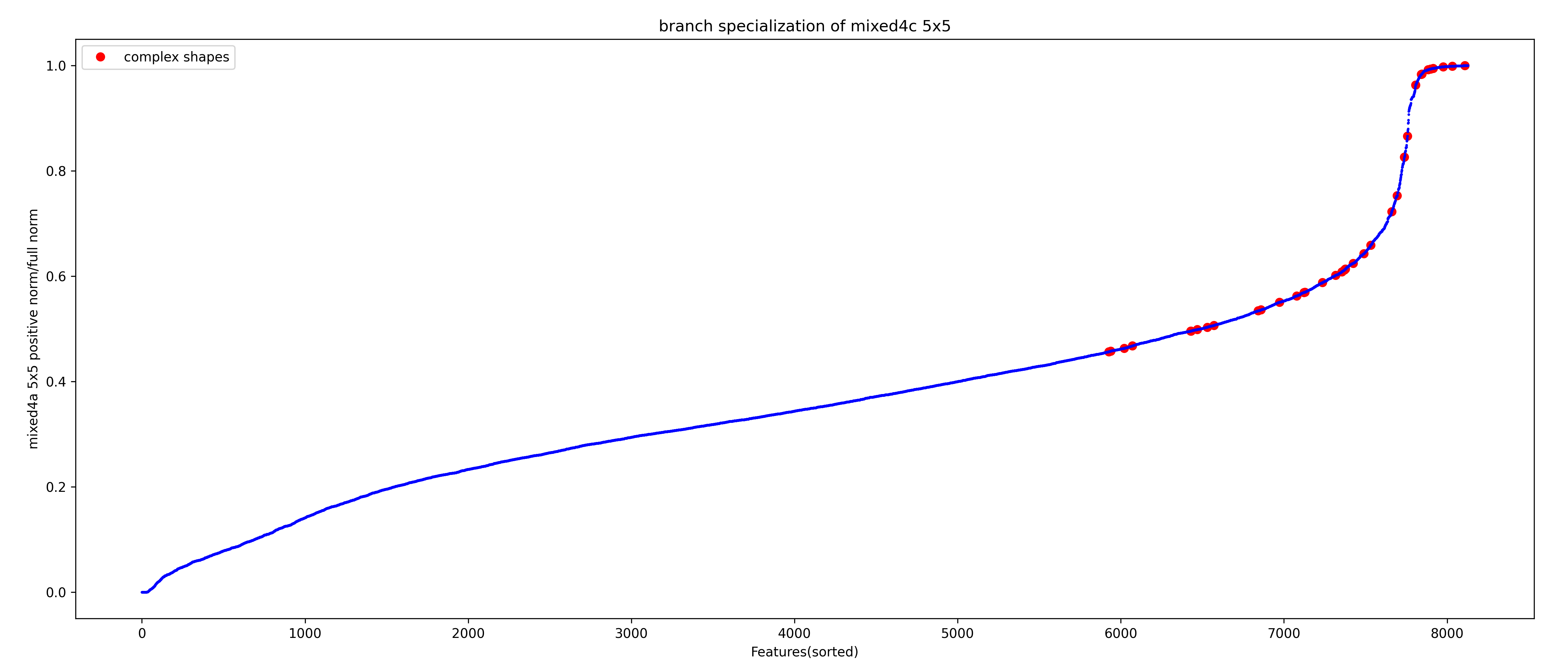}
  \caption{ How much a learned feature from the sparse autoencoder trained on all of mixed4a is represented by the neurons in branch 5x5. }
  \label{fig:graph2}
\end{figure*}
\begin{figure*}[ht]
  \centering
   
  \includegraphics[width=1\textwidth, height=1\textheight, keepaspectratio,scale=1]{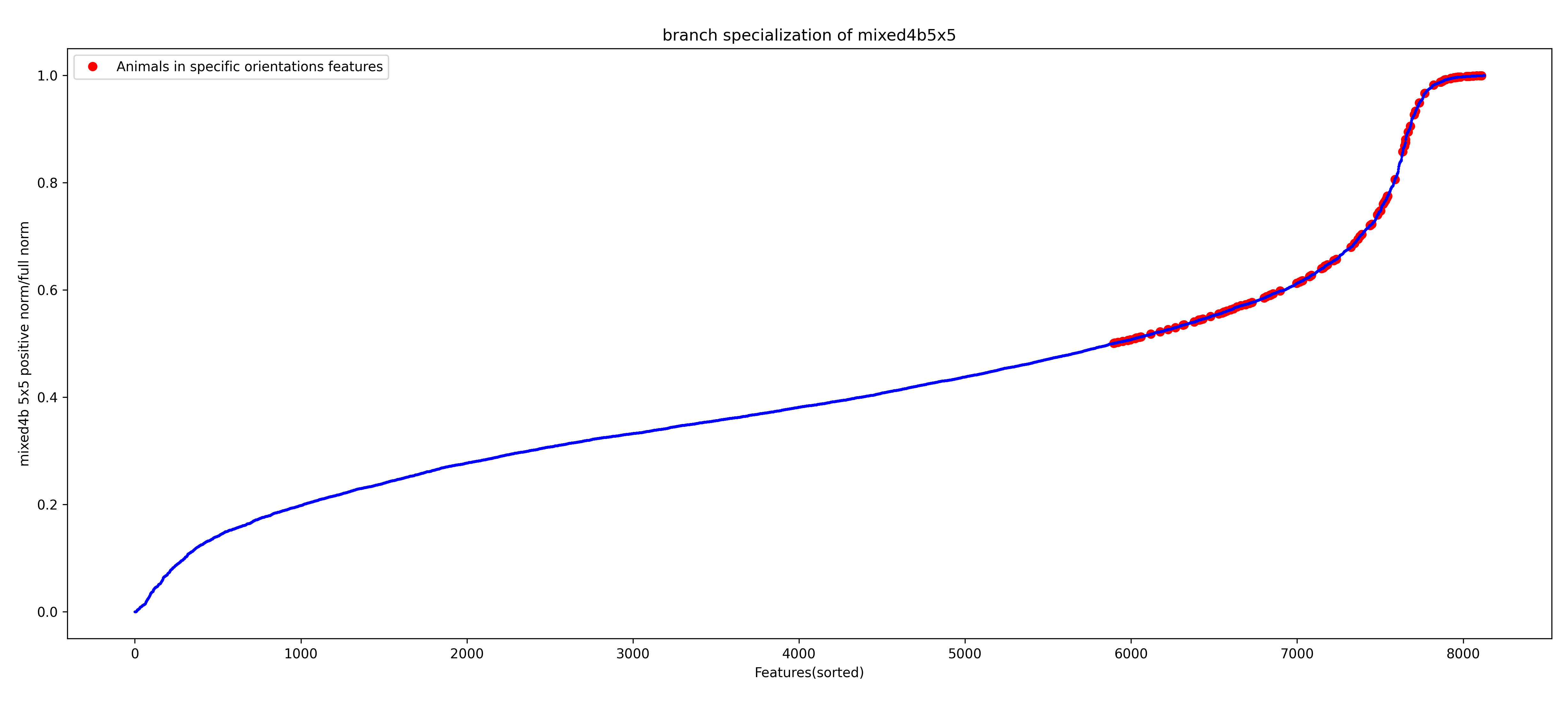}
  \caption{ How much a learned feature from the sparse autoencoder trained on all of mixed4b is represented by the neurons in branch 5x5. }
  \label{fig:graph2}
\end{figure*}
\begin{figure*}[ht]
  \centering
   
  \includegraphics[width=1\textwidth, height=1\textheight, keepaspectratio,scale=1]{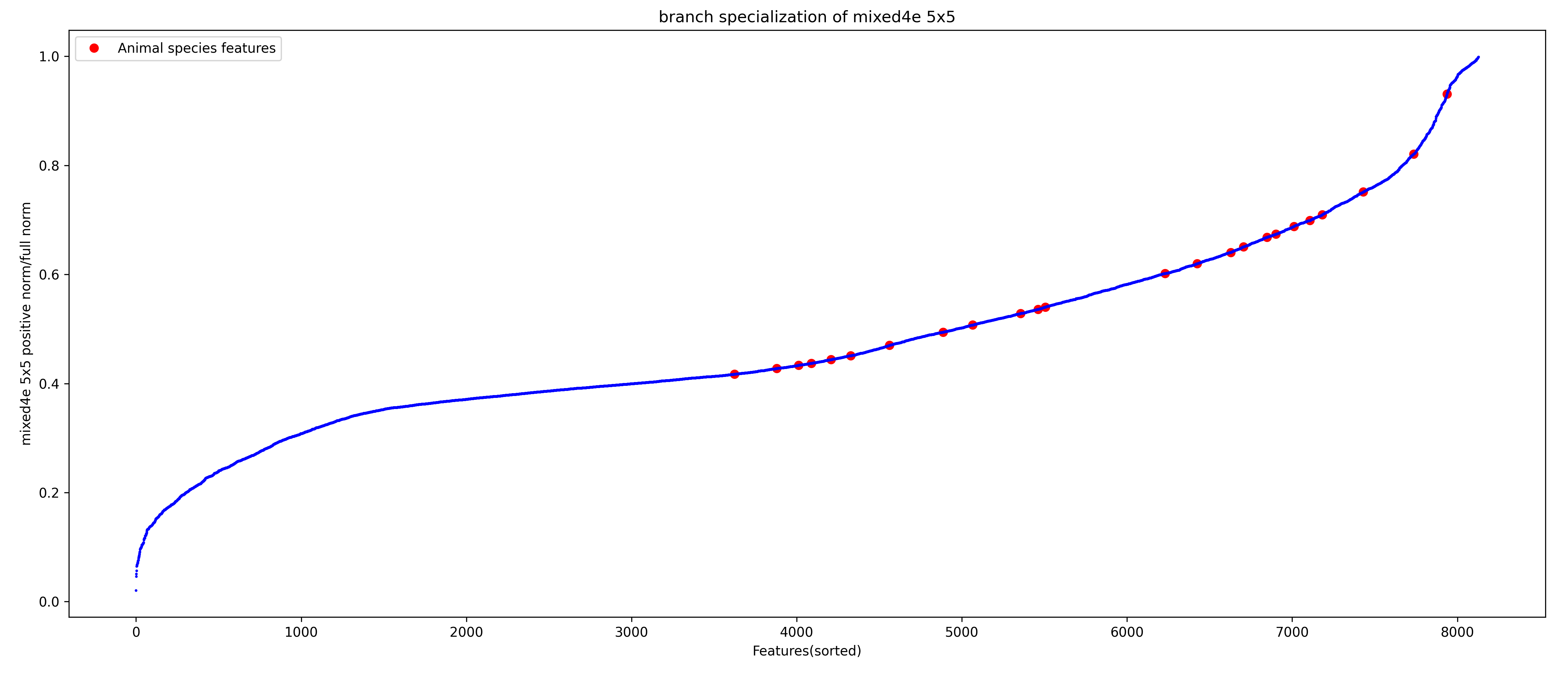}
  \caption{ How much a learned feature from the sparse autoencoder trained on all of mixed4e is represented by the neurons in branch 5x5. }
  \label{fig:graph2}
\end{figure*}


\end{document}